\documentclass{article} 
\usepackage{iclr2023_conference,times}

\usepackage{amsmath,amsfonts,bm}

\def\eqref#1{equation~\ref{#1}}

\def\1{\bm{1}}

\DeclareMathAlphabet{\mathsfit}{\encodingdefault}{\sfdefault}{m}{sl}
\SetMathAlphabet{\mathsfit}{bold}{\encodingdefault}{\sfdefault}{bx}{n}

\usepackage{hyperref}
\usepackage{url}
\usepackage{cleveref}
\usepackage{wrapfig}
\usepackage{amsthm}
\usepackage{graphicx}
\usepackage{amsmath}
\usepackage{algpseudocode}
\usepackage{float}
\usepackage{epsfig}
\usepackage{booktabs}
\usepackage{subfigure}
\newtheorem{proposition}{Proposition}

\title{Analytical Discovery of Manifold with Machine Learning}

\author{Yafei Shen\textsuperscript{1} \& Huan-Fei Ma\textsuperscript{1,}\thanks{hfma@suda.edu.cn}  \,\& Ling Yang\textsuperscript{1,}\thanks{lyang@suda.edu.cn} \\
\textsuperscript{1}School of Mathematical Sciences, Soochow University, Suzhou 215006, China\\
}

\iclrfinalcopy 
\begin{document}

\maketitle
\begin{abstract}
Understanding low-dimensional structures within high-dimensional data is crucial for visualization, interpretation, and denoising in complex datasets. Despite the advancements in manifold learning techniques, key challenges—such as limited global insight and the lack of interpretable analytical descriptions—remain unresolved.
In this work, we introduce a novel framework, GAMLA (Global Analytical Manifold Learning using Auto-encoding). GAMLA employs a two-round training process within an auto-encoding framework to derive both character and complementary representations for the underlying manifold. With the character representation, the manifold is represented by a parametric function which unfold the manifold to provide a global coordinate. While with the complementary representation, an approximate explicit manifold description is developed, offering a global and analytical representation of smooth manifolds underlying high-dimensional datasets. This enables the analytical derivation of geometric properties such as curvature and normal vectors. Moreover, we find the two representations together decompose the whole latent space and can thus characterize the local spatial structure surrounding the manifold, proving particularly effective in anomaly detection and categorization.
Through extensive experiments on benchmark datasets and real-world applications, GAMLA demonstrates its ability to achieve computational efficiency and interpretability while providing precise geometric and structural insights. This framework bridges the gap between data-driven manifold learning and analytical geometry, presenting a versatile tool for exploring the intrinsic properties of complex data sets.
\end{abstract}

\section{Introduction}
\label{Introduction}
Discovering low-dimensional structures, particularly their geometric properties, from high-dimensional data clouds enables visualization, denoising, and interpretation of complex datasets \citep{RN1201,Belkin2003LaplacianEF,Maaten2008VisualizingDU,McInnes2018UMAPUM,Luo2020DifferentiableMR}. As a result, the concept of manifold learning has attracted significant attention, leading to numerous breakthroughs over the past two decades. Various approaches have been developed, some seeking global or local linear representations, such as PCA \citep{pcabook} and random projections \citep{random_projections}. More prominent methods, such as LLE \citep{LLE2000,Donoho2003HessianEL} and ISOMAP \citep{isomap2000,Zha2007ContinuumIF}, attempt to find low-dimensional, nonlinear embedding coordinates from high-dimensional data. Despite the success of these methods in many fields, all current manifold learning techniques, to the best of our knowledge, rely on identifying neighbors for each data point. Several challenges remain unresolved with these approaches. Firstly, both the computational and space complexity are very high, especially when all pairs of shortest path distances must be computed and stored \citep{Cayton2005AlgorithmsFM,Balasubramanian2002TheIA}. Moreover, these methods are based on local information around each point, leading to a lack of global understanding of the manifold\citep{Cayton2005AlgorithmsFM}. Consequently, the learned manifold is described by local coordinates, making it difficult to rigorously calculate geometric properties or other differential indices, and providing an analytical description of the learned low-dimensional manifold is not feasible.

On the other hand, machine learning methods for exploring low-dimensional manifolds can address some of these limitations while leveraging nonlinear fitting capabilities. {Several approaches, such as manifold learning or fitting using adversarial generative networks} \citep{Yau2024,Shocher2023IdempotentGN,Khayatkhoei2018DisconnectedML,Wang2021ManifoldIF}, have emerged. These methods tend to focus on mapping from a low-dimensional manifold to a high-dimensional ambient space, rather than discovering the intrinsic properties of the manifold. Furthermore, they are often considered black-box methods, offering little in terms of analytical insight.

Beyond understanding a manifold’s global properties, identifying its local spatial structure is a crucial challenge with significant applications in anomaly detection and explanation. For example, in biomedical research, complex phenotypes arise because normal data typically reside on an underlying low-dimensional manifold, where different biological indicators exhibit intrinsic correlations. In contrast, anomalous data associated with complex phenotypes deviate from this manifold. These anomalies exist in the surrounding space of the manifold, with different deviation directions often corresponding to distinct anomaly types, while the magnitude of deviation typically reflects the severity of the anomaly. Capturing the local spatial structure of the manifold can provide deeper insights into the nature of these deviations, thereby improving anomaly interpretation. Hence, developing a manifold learning framework that also captures local spatial structures is of paramount importance.

In this paper, we address a fundamental question: \textbf{how can we provide a global analytical description of an underlying manifold with smooth geometric structure and characterize its local spatial properties} from high-dimensional data clouds? To this end, we propose a novel framework called GAMLA (Global Analytical Manifold Learning using Auto-encoding). Given a large dataset in a high-dimensional space, $\bm{x}_i \in \mathbb{R}^n$, sampled from an underlying $m$-dimensional smooth manifold $\mathcal{M}$, our approach seeks to learn the structure of $\mathcal{M}$ by integrating multiple perspectives. Specifically, we aim to construct an embedding map $G(\bm{x}): \mathbb{R}^n \to \mathbb{R}^m$ that represents the manifold in the $m$-dimensional latent space while simultaneously deriving a general analytical formulation $R(\bm{x}) = \bm{0}$, where $R: \mathbb{R}^n \to \mathbb{R}^{n-m}$, to explicitly describe the manifold in the original space. Furthermore, our framework extends beyond the manifold itself to characterize the local spatial structure around $\mathcal{M}$, offering new opportunities for understanding and interpreting the data, including applications such as anomaly detection and outlier explanation.

The proposed GAMLA framework provides a global analytical description, making it a white-box method that not only characterizes the compact and bounded manifold itself but also captures the structure of the surrounding space. Due to GAMLA's global properties, it is possible to establish a global coordinate chart, unlike piecewise approaches, thus offering a unified and consistent representation for the manifold. The analytical formulation allows for the calculation of differential geometric properties of the manifold, such as curvatures and normal vectors directly. Moreover, by eliminating the need to identify nearest neighbors, the computational cost of the learning is significantly reduced. Additionally, the function $R(x_1, x_2, \dots, x_n) =\bm{0}$ is more general than existing machine learning frameworks, which typically learn functions in the form $x_i = G(x_1, x_2, \dots, x_{i-1}, x_{i+1},\dots, x_{n})$ based on input-output relationships, whose existence may not always be guaranteed.

The paper is organized as follows: in the Methodology section, we will introduce the preliminaries, the framework, and the theoretical analysis for the proposed method. In the Results section, we will demonstrate the effectiveness of the proposed method and show the merits with both benchmark data sets as well as real world data applications. Several issues closely related to the success of the method and the further analysis of the theory will be discussed in details in the Discussions section.

\section{Methodology}
\label{Methodology}
In this section, we provide a detailed explanation of our methodology. First, we outline the theoretical assumptions and preliminaries for analytically describing a low-dimensional manifold embedded in high-dimensional space. Then, we introduce the principles of manifold learning using autoencoders, based on which we propose a method to derive the general form equation of the manifold. We also describe the specifics of data processing and model training.

\subsection{Assumptions and Preliminaries} Assume there exists an underlying smooth manifold $\mathcal{M}$ of dimension $m$ embedded in an Euclidean space $\mathbb{R}^n$, and further assume that $\mathcal{M}$ is compact and bounded. A point cloud $M = {\bm{x}_i \in \mathbb{R}^n, i = 1, 2, \dots, N}$ is uniformly sampled from $\mathcal{M}$, with a sufficiently large number of points, $N$.

Intuitively, to analytically describe the low-dimensional manifold embedded in the high-dimensional space, independent constraint functions provide the most general form. The core idea of the proposed method to discover the analytical description of an $m$-dimensional manifold embedded in $n$-dimensional space involves finding $n - m$ independent constraint functions, aided by machine learning methods. To this, we design a two-round training process to obtain both the character representation for the manifold projection as well as the complementary representation for the local spatial structure surrounding the manifold.

\subsection{First training with character representation}
\begin{figure}[h]
	\centering
	\includegraphics[width=0.6\textwidth]{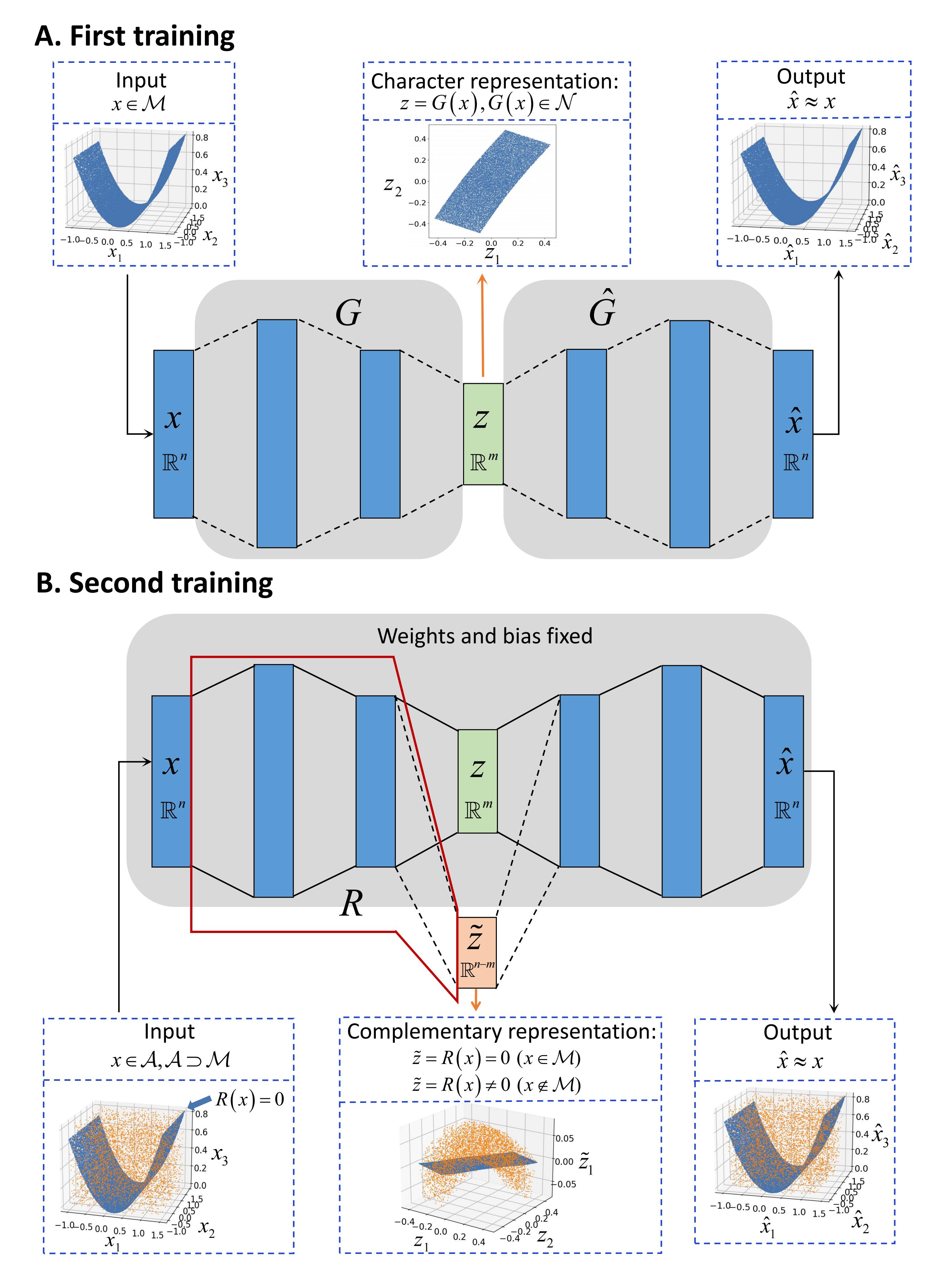}
	\caption{Illustration of GAMLA manifold learning. The training process consists of two rounds. In the first training round, the model is trained using sample points on the manifold $\mathcal{M}$. After completing this training, all weights and biases are fixed. In the second training round, only the weights and biases corresponding to the newly added nodes in the bottleneck layer are trained, using a training set comprising sample points uniformly collected from a hyperrectangle set $\mathcal{A}$ that contains the manifold $\mathcal{M}$.}\label{fig1}
\end{figure}

An autoencoder is a machine learning framework designed for dimensionality reduction in latent space \citep{autoencoder}. Typically, an autoencoder consists of a feed-forward neural network with $2k + 1$ layers, including input and output layers (Fig.\ref{fig1}A). The encoder compresses the input data into the coding space from  the $0$th to the $k$th layer, while the decoder reconstructs the compressed data back into the original space from the $k$th to the $2k$th layers. The $k$th layer, known as the bottleneck layer, represents the low-dimensional coding space.

For the first round of training, the bottleneck layer’s dimensionality is set to the intrinsic dimension of the target manifold, denoted as $m$. The dimensions of the encoder and decoder layers are typically symmetric, i.e., $n_i = n_{2k-i}$, where $n_i$ is the dimension of the $i$th layer. The dimensions of input and output layers equal to the ambient dimension of the input dataset, denoted as $n$. Additionally, we assume that the dimension of each hidden layer is not less than the ambient dimension, i.e., $n_i\geq n$.

Denote the input data as $\bm{x} \in \mathbb{R}^n$ and its respective output as $\hat{\bm{x}} \in \mathbb{R}^n$. To characterize the underlying manifold $\mathcal{M}$ from the sampled point cloud $M$, the autoencoder is trained to minimize the reconstruction error of the underlying manifold $\mathcal{M}$, with the loss function $\mathcal{L}_1$ defined as:
\begin{equation}
	\mathcal{L}_1=\sum_{\bm{x}\in M}\|\bm{x}-\hat{\bm{x}}\|,
\end{equation}
where $\|\cdot\|$ denotes the $L_2$ norm in $\mathbb{R}^n$.

After the first round of training, any sampled point $\bm{x} \in \mathcal{M}$ is reconstructed as $\hat{\bm{x}}$, satisfying $\|\hat{\bm{x}} - \bm{x}\|_2^2 < \xi$  where $\xi$ is a threshold value. Letting the nodes in the bottleneck layer be denoted by $\bm{z} \in \mathbb{R}^m$, the encoder establishes a map $\bm{z} = G(\bm{x})$ which fully characterizes the manifold $\mathcal{M}$. Thus, the space formed by $\bm{z}$ is referred to as the \textbf{character space}. Consequently, after this training phase, the encoder map $\bm{z} = G(\bm{x})$ functions as an embedding from the original $n$-dimensional space to the $m$-dimensional character space. Meanwhile, the decoder defines a parametric equation
\begin{equation}\label{parametric equation}
	\hat{\bm{x}} = \hat{G}(\bm{z}),
\end{equation}
which represents the manifold $\mathcal{M}$ in terms of the coordinates in the character space. Thus, the underlying manifold is considered as fully reconstructed with respect to the character space, and $\bm{z} = G(\bm{x})$ is referred as the \textbf{character representation} for the manifold $\mathcal{M}$.

\subsection{Second training with complementary representation}
After the first training, to further derive the general equation describing the manifold $\mathcal{M}$ in the original space, we add $n - m$ nodes to the bottleneck layer, forming a complementary architecture that aligns the bottleneck layer with the original space dimension (Fig.\ref{fig1}B). In this second training phase, only the weights related to the newly added bottleneck nodes are trained, while the weights obtained from the first training are fixed. The goal of this phase is to minimize the loss function $\mathcal{L}_2$, which measures the reconstruction error for the entire ambient space:
\begin{equation}
	\mathcal{L}_2=\sum_{\bm{x}\in A}\|\bm{x}-\hat{\bm{x}}\|,
\end{equation}
where $A$ is a point cloud sampled uniformly from the ambient set $\mathcal{A}$, with $\mathcal{A} \supset \mathcal{M}$. In this study, we define $\mathcal{A}$ as a hyperrectangle set containing $\mathcal{M}$.

After the second training, let the map from the input to the newly added nodes be $\tilde{\bm{z}} = R(\bm{x}) \in \mathbb{R}^{n-m}$. For such a map $R$, we have the following proposition which implies that $\mathcal{M}$ is fully reconstructed by the kernel of $R$.

\begin{proposition}\label{prop1} If the underlying manifold $\mathcal{M}$ is fully reconstructed in the first training, the map $\tilde{\bm{z}} = R(\bm{x}) \in \mathbb{R}^{n-m}$ generated from the second training satisfies $R(\bm{x}) = \bm{0}$ for all $\bm{x} \in \mathcal{M}$. \end{proposition}

In light of Proposition \ref{prop1}, the equation
\begin{equation}\label{general equation}
	R(\bm{x}) = \bm{0},
\end{equation}
provides \(n-m\) independent constraints for the \(m\)-dimensional manifold \(\mathcal{M}\) embedded in the \(n\)-dimensional original space. Therefore, after the second round of training, Eq. \ref{general equation} offers a general equation describing the manifold \(\mathcal{M}\) in the original space.

It is important to note that the newly added nodes $\tilde{\bm{z}} \in \mathbb{R}^{n-m}$, together with $\bm{z} \in \mathbb{R}^m$ from the first training round, form the entire \(n\)-dimensional autoencoder latent space \([\bm{z}, \tilde{\bm{z}}] \in \mathbb{R}^n\), which is aligned with the original \(n\)-dimensional space. Thus, the space spanned by \(\tilde{\bm{z}}\) is referred to as the \textbf{complementary space} which complements the character space spanned by $\bm{z}$, and $\tilde{\bm{z}} = R(\bm{x})$ is termed the \textbf{complementary representation}. In contrast to the parametric representation in Eq. \ref{parametric equation}, which aligns with other embedding-based approaches in the literature \citep{Yau2024}, the general equation derived from the complementary representation represents a novel contribution. It is the first equation to globally characterize the manifold's geometry within the original space, providing a powerful tool for describing the underlying manifold’s geometric properties. Furthermore, since \(R(\bm{x}) = \bm{0}\) is a set of explicit analytical equations, this approach is both interpretable and differentiable, facilitating the derivation of differential geometric properties. In some cases, the precise formulation of the manifold can be further reconstructed, as shown in the results section.

Moreover, the latent space formed by \([\bm{z}, \tilde{\bm{z}}]\) not only provides two distinct descriptions of the manifold but also allows for the characterization of the local spatial structure surrounding the manifold. Specifically, for a point \(\bm{x}\) lying outside the manifold \(\mathcal{M}\), i.e., \(\bm{x} \notin \mathcal{M}\), the decoder output \(\hat{\bm{x}} = \hat{G}(\bm{z})\) gives the projection of \(\bm{x}\) onto the manifold, while the value of \(\tilde{\bm{z}} = R(\bm{x})\) describes how \(\bm{x}\) deviates from the manifold. This is especially useful for anomaly detection and categorization, as discussed further in the results section.

From an efficiency standpoint, this approach is computationally efficient since it eliminates the need to determine nearest neighbors or compute pairwise distances. As a result, the two-round training framework is named \textbf{GAMLA} (Global Analytical Manifold Learning using Auto-encoding). 

\section{Results}
In this section, we present the key outcomes of the GAMLA framework, organized into three main aspects. First, we demonstrate the complementary representation using analytical mathematical formulations that accurately capture its intrinsic geometry. This reconstruction lays the foundation for precise representation and understanding of the manifold's structure. Then, we illustrate the learning of the manifold structure by effectively unfolding the manifold in the character space, thereby establishing a global coordinate chart. This approach enables coherent navigation across the manifold and facilitates downstream applications requiring global parametrization. Furthermore, we explore the learning ability of GAMLA framework for the local spatial structures outside the manifold. This capability is critical for identifying and explaining deviations from the manifold and can be applied to tasks that require an understanding of the intrinsic structure of the main data as well as details of deviations in other data, such as anomaly categorization and outlier explanation. The results are illustrated with both benchmark data sets and real data applications.

\subsection{Analytical Mathematical Formulation Reconstruction and Differential Geometric Property}
\begin{figure}[h]
	\centering
	\includegraphics[width=0.99\textwidth]{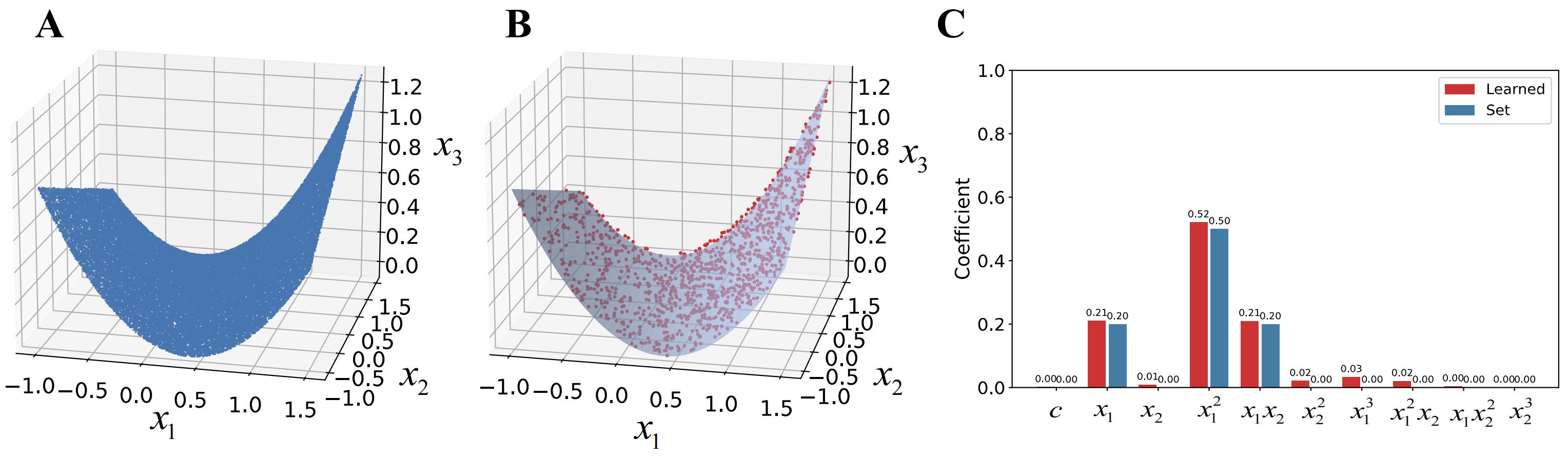}
	\caption{Estimation results of the mathematical expression for manifold with a global explicit function $x_3=-0.2x_1+0.5x_1^2+0.2x_1x_2$. (\emph{A}) Scatter plot of quadric surface. (\emph{B}) The red data points selected through the approximation expression and the ground truth manifold depicted in gray. (\emph{C}) Comparison of the coefficients of the Taylor expansion of the estimated expressions for the quadric surface with the true expressions.}\label{fig2}
\end{figure}

We begin with a typical manifold, a quadric surface, which can be explicitly expressed as:
\begin{equation}\label{manifold1}
	x_3=-0.2x_1+0.5x_1^2+0.2x_1x_2,
\end{equation}
where a data cloud of $40000$ points is uniformly sampled within the region $x_1 \in (-1, 1.5), x_2 \in (-1, 1.5)$, as shown in Fig.\ref{fig2}A. Using a $5$ layer GAMLA with a structure size of $(3,3,2,3,3)$ and after two rounds of training we obtain the general function for the manifold as
\begin{equation}\label{generalfunction1}
	R_1(x_1,x_2,x_3)=\sum_{i=1}^{3}\omega_{i}\tanh(\sum_{j=1}^{3}a_{ij}x_j+c_i)+d,
\end{equation}
with $d = 0.6553037$ and other coefficients listed in Table \ref{table1}. To validate the correctness of $R_1$, we generate $10^5$ uniformly distributed points and observe that nearly all points satisfying $|R_1(x_1,x_2,x_3)|<\varepsilon_1$ with $\varepsilon_1=0.001$ lie on the manifold, as shown in Fig.\ref{fig2}B, confirming that $R_1$ accurately represents the manifold.

To further confirm the analytical property of $R_1$, we express the implicit function derived from $R_1(x_1,x_2,x_3)=0$ as $x_3=f(x_1,x_2)$ and perform a Taylor expansion up to the third order at the origin and the expansion coefficients are depicted in Fig.\ref{fig2}C. After setting a non-zero threshold of $0.03$, the Taylor expansion of Eq.\ref{generalfunction1} yields
\begin{equation}\label{taylor1}
	x_3=-0.21x_1+0.52x_1^2+0.21x_1x_2,
\end{equation}
which closely approximates the ground truth in Eq.\ref{manifold1}.

\begin{table}[H]
	\centering
	\caption{Learned coefficients for Eq.\ref{generalfunction1}}\label{table1}
	\begin{tabular}{cccc}
		\toprule
		Coefficients & $i=1$ & $i=2$ & $i=3$ \\
		\midrule
		$a_{i1}$ & 0.2093325 & 0.3301388 & -0.5808623 \\
		$a_{i2}$ & 0.04949709 & 0.07136403 & -0.11274612 \\
		$a_{i3}$ & -0.15945598 & -0.17968029 & -0.15649709\\
		$c_{i}$ & 1.0170152 & -0.16471033 & -1.3647013\\
		$\omega_{i}$ & 0.6709025 & 0.26675373 & 1.2869824\\
		\bottomrule
	\end{tabular}
\end{table}

\begin{figure}[h]
	\centering
	\includegraphics[width=0.99\textwidth]{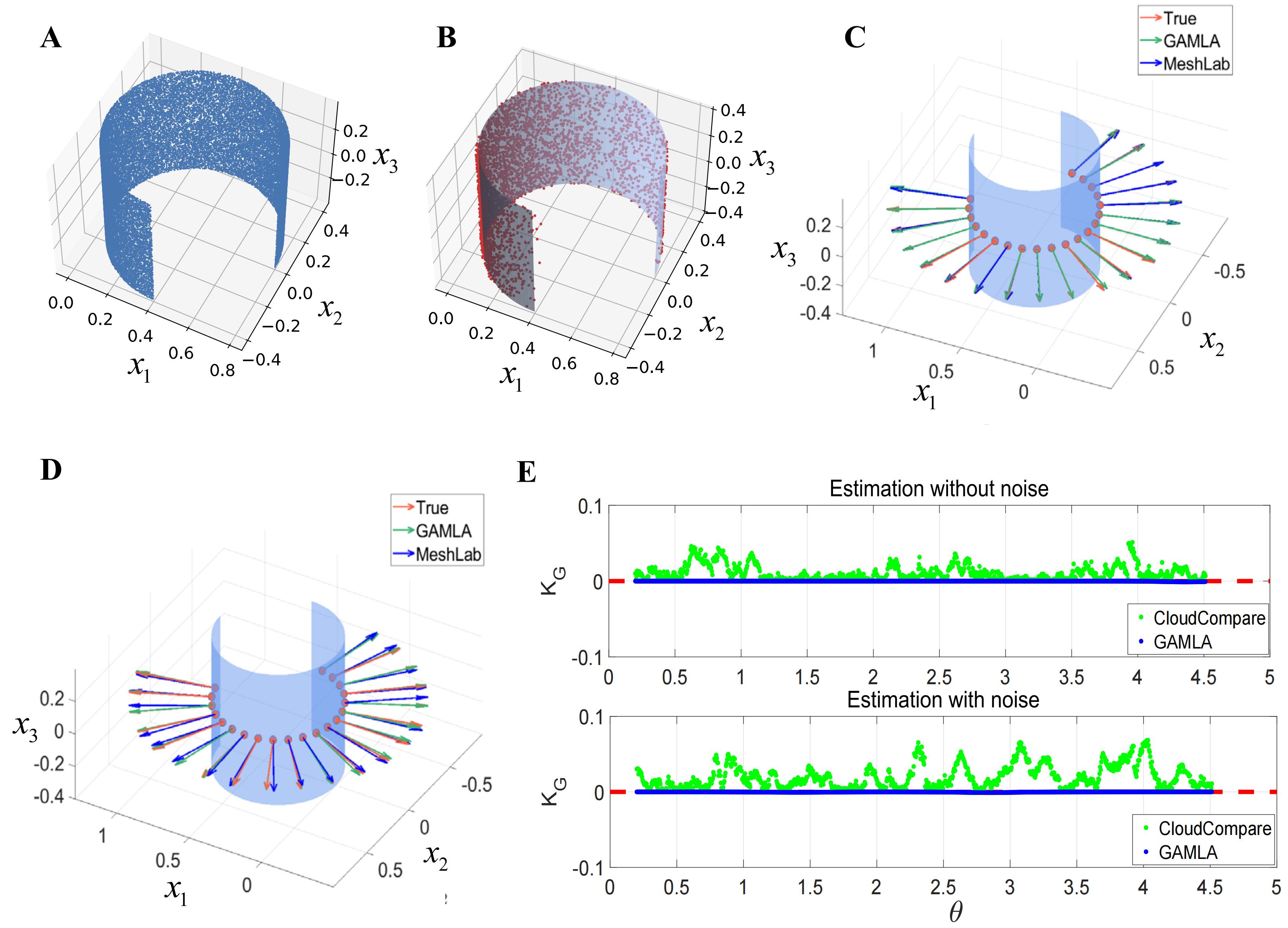}
	\caption{Estimation results of the mathematical expression for manifolds without a global explicit function. (\emph{A}) Scatter plot of three-quarter cylinder. (\emph{B}) The red data points selected through the approximation expression $|R_2|<\epsilon$ and the true manifold depicted in gray. (\emph{C}) The normal vectors calculated using the GAMLA analytically and using MeshLab numerically. (\emph{D}) The same as (\emph{C}) but with noisy sampling data cloud.  (\emph{E}) The Gaussian curvature (ground truth: $K_G=0$) where $x_3=0$ on the three-quarter cylinder calculated using GAMLA analytically and using CloudCompare software numerically from data cloud without noise (up) and with noise (down) respectively.}\label{fig3}
\end{figure}

Next, we consider a more challenging scenario where the manifold is a three-quarter cylinder that satisfies:
\begin{equation}
	\label{manifold2}
	\left\{
	\begin{aligned}
		& x_1=0.4\cos(\theta)+0.4\\
		& x_2=0.4\sin(\theta)\\
		& x_3=l,
	\end{aligned}
	\right.
\end{equation}
where $l \in (-0.4,0.4)$ and $\theta \in (0,1.5\pi)$, with $40000$ data points uniformly sampled, as shown in Fig.\ref{fig3}A.  Using a GAMLA with a structure of $(3,6,6,2,6,6,3)$, we obtain the general function $R_2(x_1,x_2,x_3)=0$. To validate $R_2$, we firstly generate $10^5$ data points uniformly in the cubic $[0,0.8]\times[-0.4,0.4]\times[-0.4,0.4]$ and depict all the points satisfying $|R_2(x_1,x_2,x_3)|<\varepsilon$, with $\varepsilon=0.005$, as shown in Fig.\ref{fig3}B. This confirms that $R_2(x_1,x_2,x_3)=0$ accurately represents the manifold.

Unlike Eq. \ref{manifold1}, Eq. \ref{manifold2} cannot be expressed as an explicit function in any dimension. Thus, verifying the analytical form through Taylor expansion is not possible. To demonstrate the utility of the analytical form  $R_2$, we carry out two tasks to show that $R_2$ can be used to calculate differential properties directly.
To this, we compute the normal vector as well as Gaussian curvature directly from the formulation $R_2$ analytically around a series of points with details appended in SI. Fig.\ref{fig3}C shows that normal vectors calculated from $R_2$ align closely with the ground truth.  Additionally, the Gaussian curvature along the surface at $x_3=0$, as shown in Fig.\ref{fig3}E, matches well with the true value $K_G=0$.

Although it is also possible to estimate normal vectors and Gaussian curvature numerically from point clouds directly, the analytical approach is more powerful, especially with unevenly distributed or noisy data.
We compare these analytical results with numerical estimates obtained using the software Meshlab and CloudCompare \citep{Cignoni2008MeshLabAO}(as shown in Fig.\ref{fig3}C, E). It is clear that when the data is noiseless, the normal vectors calculated from the analytical form $R_2$ and estimated numerically by Meshlab are both close to the ground truth while the Gaussian curvatures calculated by $R_2$ is much better than the numerical estimation by CloudCompare. For noisy data, the GAMLA framework, which reconstructs the manifold using global information, significantly outperforms local numerical methods, as illustrated in Fig.\ref{fig3}D, E.

These examples confirm that our method can provide an accurate mathematical expression for smooth, compact, and bounded manifolds. When the manifold can be globally represented as an explicit function, our results are consistent with Taylor series approximations. For manifolds that cannot be globally represented explicitly, our approach still produces accurate approximations, and the results do not depend on the choice of basis, as they are learned automatically by the GAMLA framework.

\subsection{Manifold Unfolding and Global Coordinate Chart}
In this section, using character representation, we illustrate the manifold unfolding from high-dimensional original space to low-dimensional character space.
Specifically, with such an unfolding, we demonstrate that the GAMLA framework can be used to provide a global coordinate chart for the underlying manifold.

\begin{figure}[h]
	\centering
	\includegraphics[width=0.6\textwidth]{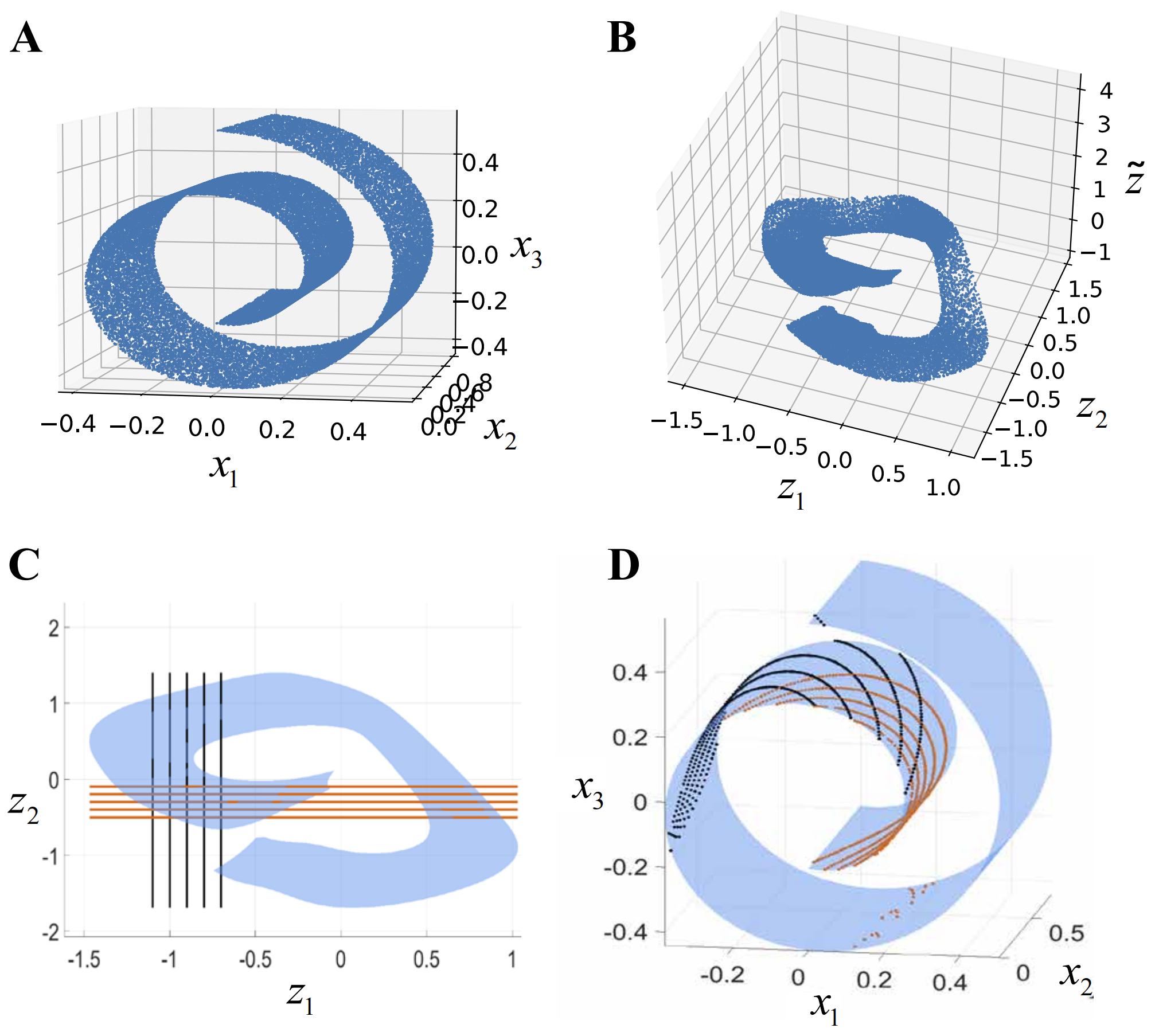}
	\caption{Unfolding and global coordinate chart acquisition results for the Swiss Roll manifold. (\emph{A}) Scatter plot of the Swiss Roll manifold. (\emph{B}) The unfolding of the Swiss Roll manifold in latent space. (\emph{C}) and (\emph{D}) The axes in the character space corresponding to the global coordinate chart in the original space.}\label{fig4}
\end{figure}

We take the renowned Swiss Roll manifold as a benchmark which is defined by the parametric equations:
\begin{equation}
	\label{manifold3}
	\left\{
	\begin{aligned}
		& x_1 = 0.04t\cos(t)\\
		& x_2=l\\
		& x_3=0.04t\sin(t).
	\end{aligned}
	\right.
\end{equation}
where $t=1.5\pi(1+2r)$, $r \in (0,1)$ and $l \in (0,0.8)$.  To reconstruct this manifold, we design the GAMLA framework in the first round of training using $9$ fully connected layers with size $(3,24,12,6,2,6,12,24,3)$.  The training dataset consists of $40000$ sampled points as shown in Fig.\ref{fig4}A.

In the second round of training, we add one node to the bottleneck layer. After two rounds of training, the output of each data point in the bottleneck layer is denoted as $\bm{z}=(z_1,z_2,\tilde{z})$ where $z_1,z_2$ correspond to the original two nodes while $\tilde{z}$ represents the added node. Therefore,  $\bm{z}$ can be considered as new coordinates in the latent space obtained by GAMLA. Fig.\ref{fig4}B shows the distribution of the $\bm{z}$ as an image of the sampled data in the latent space where it is clearly shown that these data points satisfying $\tilde{z}\approx 0$, as predicted by Proposition \ref{prop1}. This confirms that GAMLA has successfully unfolded the 3-dimensional Swiss Roll from the original space to the 2-dimensional character space. Importantly, the axes in the character space provide a new global coordinate system $(z_1,z_2)$. To illustrate this, we show several grid lines in the character space and project them back into the original space (Fig.\ref{fig4}C and Fig.\ref{fig4}D). These results demonstrate that the learned character space provides a global coordinate chart for the manifold, in contrast to the piecewise local charts used in other methods.

\begin{figure}[h]
	\centering
	\includegraphics[width=0.6\textwidth]{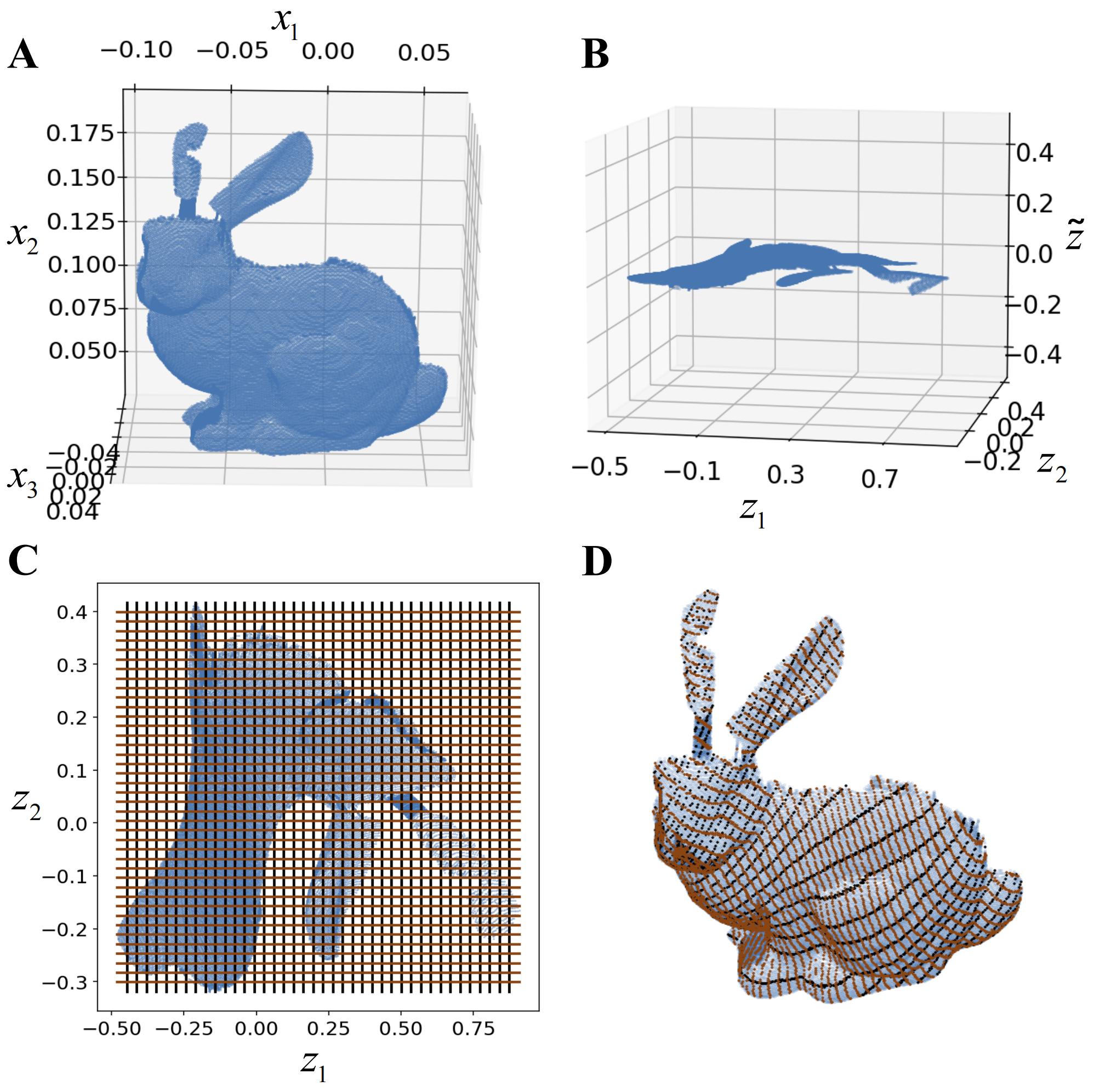}
	\caption{Unfolding and global coordinate chart acquisition results for $3$D point cloud dataset. (\emph{A}) Scatter plot of 3D point cloud dataset collected from the Stanford Bunny model. (\emph{B}) The unfolding of the $3$D point cloud dataset in latent space. (\emph{C}) and (\emph{D}) illustrate that the grid lines in the character space form quad mesh partitions of the surface where the 3D point cloud dataset lies.}\label{fig5}
\end{figure}

To further validate the applicability of the GAMLA framework, we apply it to a real dataset: the \href{https://graphics.stanford.edu/data/3Dscanrep/}{Stanford Bunny model} from Stanford Computer Graphics Laboratory. This dataset contains $48866$ data points collected from this $3$D scanning model (as shown in Fig. \ref{fig5}A), distributed on a manifold with an intrinsic dimension of $2$. The results shown in Fig.\ref{fig5}B further validate the effectiveness in manifold unfolding and global coordinate chart acquisition in this real data set. Notably, lines parallel to the axes in the character space can be used for quad mesh partitioning of the surface, where the 3D point cloud dataset resides (Fig.\ref{fig5}D). This demonstrates the potential application of GAMLA for quad mesh partitioning in surface analysis.

\subsection{Local Spatial Structure and Anomaly Categorization}
\begin{figure}[h]
	\centering
	\includegraphics[width=0.99\textwidth]{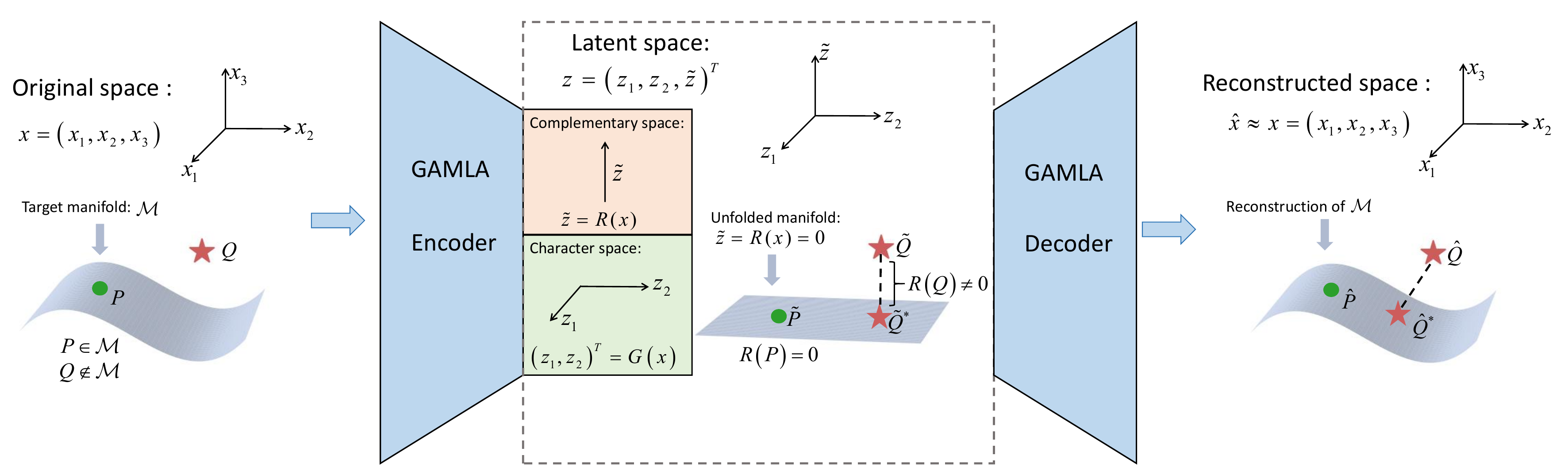}
	\caption{Schematic illustration of the spatial decomposition of the latent space in GAMLA. $P$ and $Q$ are points located on and off the target manifold $\mathcal{M}$ in the original space, respectively. In the latent space of GAMLA, the corresponding points of $P$ and $Q$ are denoted as $\tilde{P}$ and $\tilde{Q}$. The complementary representation of points on the manifold $\mathcal{M}$ always satisfies $\tilde{\bm{z}}=R(\bm{x})=0$, and $\tilde{Q}^*$ represents the projection of $\tilde{Q}$ onto the unfolded manifold. The character representation of $Q$ consists of the coordinates of $\tilde{Q}$ and $\tilde{Q}^*$ in the character space, while the complementary representation of $Q$ describes the deviation of $\tilde{Q}$ from the unfolded manifold. Through the action of the decoder, all points in the original space are fully reconstructed. In addition, the decoder output of $\tilde{Q}^*$ is $\hat{Q}^*$, which can be regarded as the projection of $Q$ onto the manifold $\mathcal{M}$.}\label{fig11}
\end{figure}

From a space decomposition perspective, the latent space of GAMLA spanned by $[\bm{z},\tilde{\bm{z}}]$ is decomposed into the sum of two subspaces: the character space spanned by $\bm{z}=G(\bm{x})$ and its complementary space spanned by $\tilde{\bm{z}}=R(\bm{x})$. For the points on the underlying manifold $\mathcal{M}$, these two subspaces provide two distinct representations: while $\hat{G}(\bm{z})$ provides a parametric function in the character space, $R(\bm{x})=0$ describes the manifold with a general function. As a matter of fact, these two subspaces together also provide a way to characterize the local spatial structure surrounding the underlying manifold. Specifically, for a point lying outside the manifold $\mathcal{M}$, i.e., $\bm{x}\notin\mathcal{M}$, the output using the character representation provides the projection of $\bm{x}$ onto the manifold while the complementary representation $\tilde{z}$ describes how $\bm{x}$ deviates from the manifold, as illustrated in Fig.\ref{fig11}.

\begin{figure}[h]
	\centering
	\includegraphics[width=0.7\textwidth]{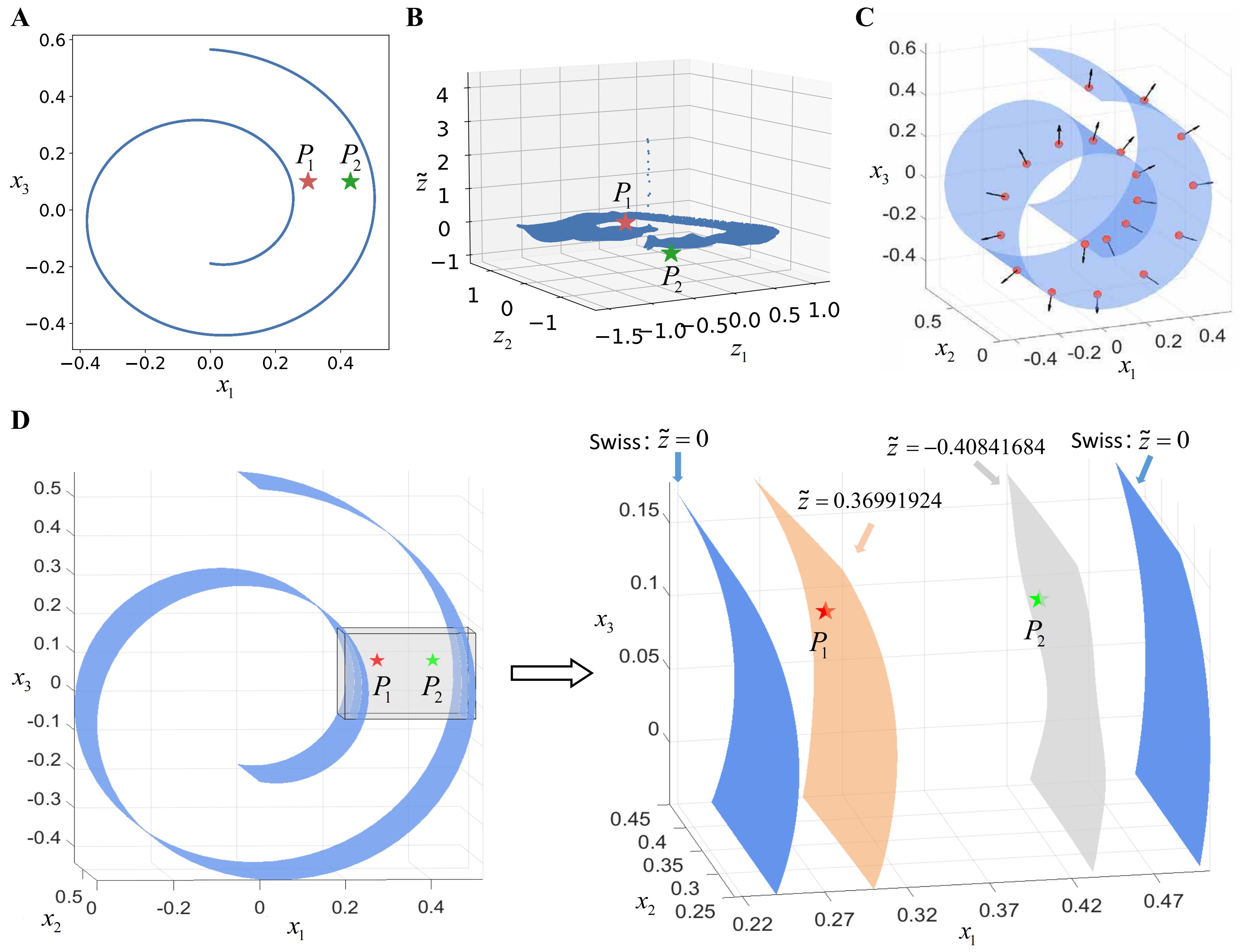}
	\caption{Results of learning the local spatial structure based on newly added bottleneck nodes. (\emph{A}) and (\emph{B}) illustrate the distribution of points outside the manifold in the latent space. (\emph{C}) The normal vectors at each point on the manifold calculated based on the estimated expression. (\emph{D}) The local spatial structure around the manifold learned by GAMLA.}\label{fig6}
\end{figure}

To further illustrate this, we still use the Swiss Roll as an example. Since the Swiss Roll is a 2-dimensional manifold embedded in 3-dimensional space, the complementary representation \(\tilde{z}\) is a scalar, with its sign indicating the orientation of the manifold and distinguishing the surrounding space. Consider two points outside the Swiss Roll: \(P_1(0.3, 0.4, 0.1)\) and \(P_2(0.43, 0.4, 0.1)\). After calculating their \(\tilde{z}\) values using the trained GAMLA framework, we visualize both points in the original and latent spaces, as shown in Fig.\ref{fig6}A and Fig.\ref{fig6}B respectively. Interestingly, although both points lie outside the manifold with nonzero \(\tilde{z}\) values, \(P_1\) (with \(\tilde{z}_1 = 0.36991924\)) lies ``above" the unfolded manifold in the latent space, while \(P_2\) (with \(\tilde{z}_2 = -0.40841684\)) lies ``below" it. This directionality is further confirmed by the normal vectors in the original space, calculated directly from \(R_3(x_1, x_2, x_3) = 0\) as shown in Fig.\ref{fig6}C.
The local spatial structure around the manifold is further illustrated in Fig.\ref{fig6}D, where the manifolds in the original space are regenerated by setting \(\tilde{z} \equiv 0\), \(\tilde{z} \equiv \tilde{z}_1\), and \(\tilde{z} \equiv \tilde{z}_2\). Notably, the manifolds corresponding to \(\tilde{z} \equiv \tilde{z}_1\) and \(\tilde{z} \equiv \tilde{z}_2\) exhibit local geometric features consistent with the Swiss Roll manifold, effectively capturing the local spatial structure surrounding it.

\begin{figure}[h]
	\centering
	\includegraphics[width=0.99\textwidth]{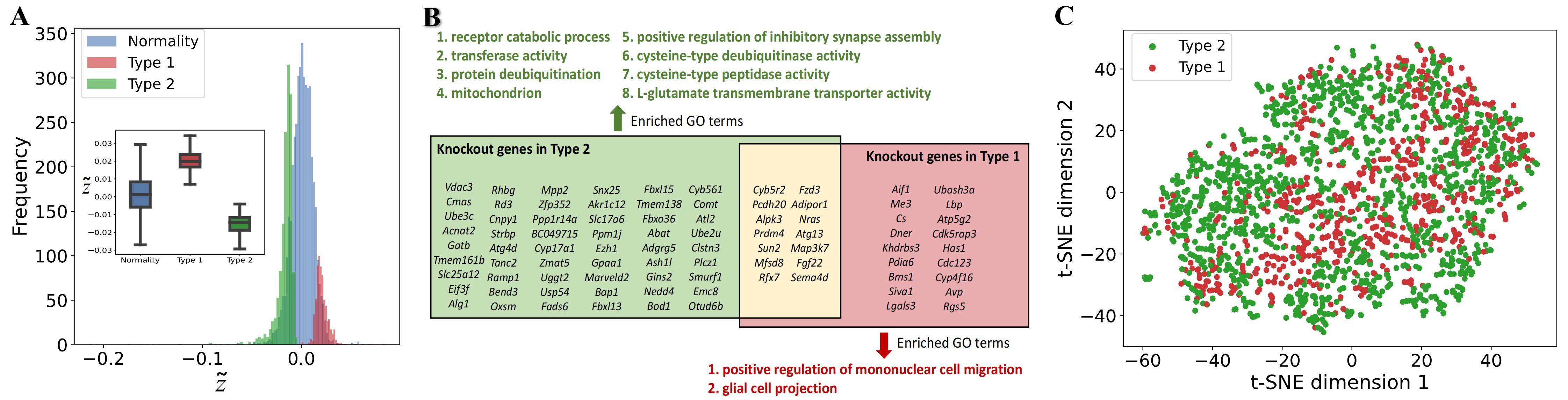}
	\caption{Application of GAMLA in the classification of outliers from single-gene knockout mice. (\emph{A}) Distribution of $\tilde{z}$ values of outliers in the latent space. (\emph{B}) GO terms enriched for knockout genes corresponding to the two groups of anomalous data using GOATOOLS. (\emph{C}) Visualization of two groups of outliers based on t-SNE.}\label{fig10}
\end{figure}

The local spatial structure learned by GAMLA indicates its ability to detect outliers and to differentiate between various types of anomalies. To this, we analyze anomalies from the single-gene knockout mouse dataset provided by the \href{https://www.mousephenotype.org/}{International Mouse Phenotyping Consortium (IMPC)}. This dataset contains nine physiological indicators for both wild-type and single-gene knockout mice, resulting in a 9-dimensional dataset. Given the inherent correlations often observed among different physiological indicators, the data points from normal mice are assumed to be distributed on an underlying manifold, which can be learned using GAMLA. In contrast, data points from single-gene knockout mice are potential outliers that deviate from this manifold.
Specifically, we use a training set of $4332$ data points from wild-type mice, all considered normal, to learn the target manifold. We determine that the intrinsic dimensionality of the target manifold is approximately $8$, leading to the representation $\bm{z} = (z_1, \dots, z_8, \tilde{z})$, where $z_i, i = 1, 2, \dots, 8$ form the character space and $\tilde{z}$ spans the complementary space. To identify anomalies, we input all $10873$ single-gene knockout mouse data points into the trained GAMLA model, identifying outliers based on their reconstruction errors. These anomalies are further analyzed using the complementary representation $\tilde{z}$ and classified into two distinct groups based on the sign of their $\tilde{z}$ values: Type 1 ($\tilde{z} > 0$) and Type 2 ($\tilde{z} < 0$), as shown in Fig.\ref{fig10}A.
To assess the biological relevance of this categorization, we conduct an analysis using the GOATOOLS package \citep{Klopfenstein2018GOATOOLSAP}, which reveals significant differences in Gene Ontology (GO) \citep{Ashburner2000GeneOT} terms enriched by the knockout genes associated with these two anomaly groups (Fig.\ref{fig10}B). This finding underscores functional distinctions between the two gene sets, further supporting the divergence between the two types of anomalies. In contrast, applying t-SNE directly to the anomalies without leveraging GAMLA fails to distinguish between the two types, as shown in Fig.\ref{fig10}C. Thus, the proposed GAMLA framework enables the discovery of significantly distinct types of anomalies in a purely data-driven manner, which cannot be achieved by existing clustering methods. Furthermore, these discoveries are validated through gene analysis, highlighting the utility and robustness of the GAMLA approach.

\subsection{Application in Images}
In the application of manifold learning to images, it is widely accepted that images of a specific pattern are assumed to lie on a particular manifold in high-dimensional space \citep{isomap2000,Khayatkhoei2018DisconnectedML}. This implies that interpolations in the character space of GAMLA can yield a continuous gradient process between two images.
To explore this idea, we conduct a series of experiments on image datasets, aiming to expand the scope of applications for our approach. These experiments demonstrate how GAMLA can generate smooth transitions between images, effectively capturing the underlying structure of the image manifold.
\begin{figure}[ht]
	\centering
	\includegraphics[width=0.99\textwidth]{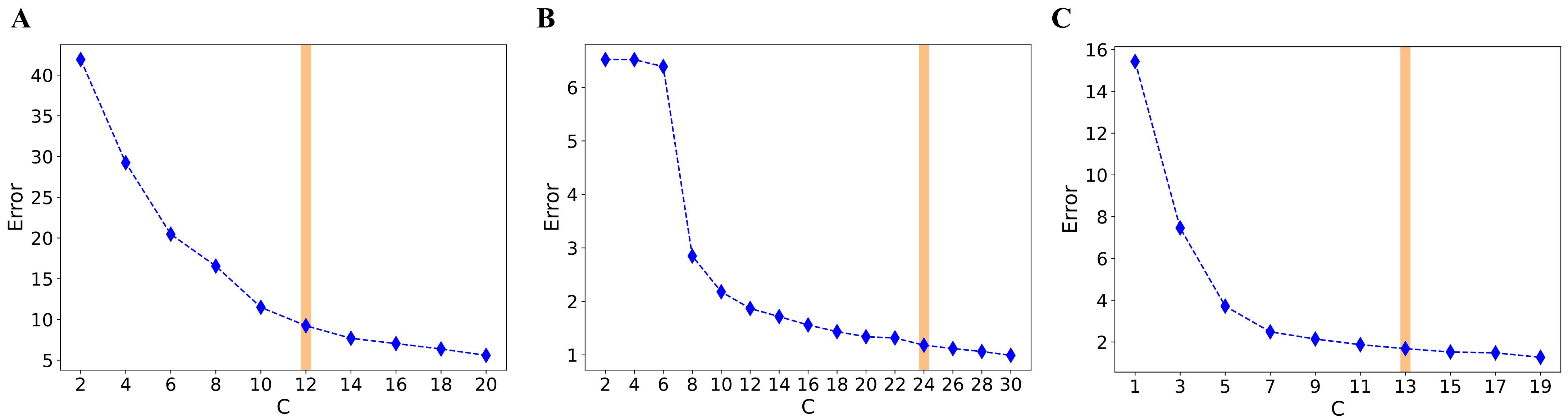}
	\caption{Estimation of the intrinsic dimension of the manifold where the image dataset lies. (\emph{A}) Handwritten ``2'' from the MNIST dataset; (\emph{B}) ``Gray Face'' dataset; (\emph{C}) The handwritten ``1'' from the augmented MNIST dataset.}\label{fig7}
\end{figure}
Firstly, we consider the MNIST dataset for handwritten digits. We select $5958$ images of the handwritten ``2" from the dataset, each with a pixel size of $28 \times 28$, as the training dataset. Generally, each image of a specific digit is assumed to be a data point lying on a specific manifold in the high dimensional space with an ambient dimension of $784$. To determine the intrinsic dimension of this unknown manifold, we first carry out a dimensionality test. Specifically, we fix the size of the GAMLA framework and gradually increase the size of the bottleneck layer. For each bottleneck layer size, we run the first round of training $10$ times and record the total reconstruction error for all the images in the training set as shown in Fig.\ref{fig7}A. Since the reconstruction error reflects the reconstruction performance of GAMLA, the inflection point of the error curve indicates further increasing the size of bottleneck layer does not lead to significant decrease of training error. Therefore, we consider the point around the elbow of the error curve as the intrinsic dimension and assume that the first round of training with such an intrinsic dimension has achieved complete reconstruction of the manifold. Fig.\ref{fig7}A illustrates that the intrinsic dimension of the manifold where the images of handwritten ``2" lie is approximately $12$. We then set the bottleneck layer dimension of the autoencoder to $12$ to reconstruct all images.

\begin{figure}[h]
	\centering
	\includegraphics[width=0.99\textwidth]{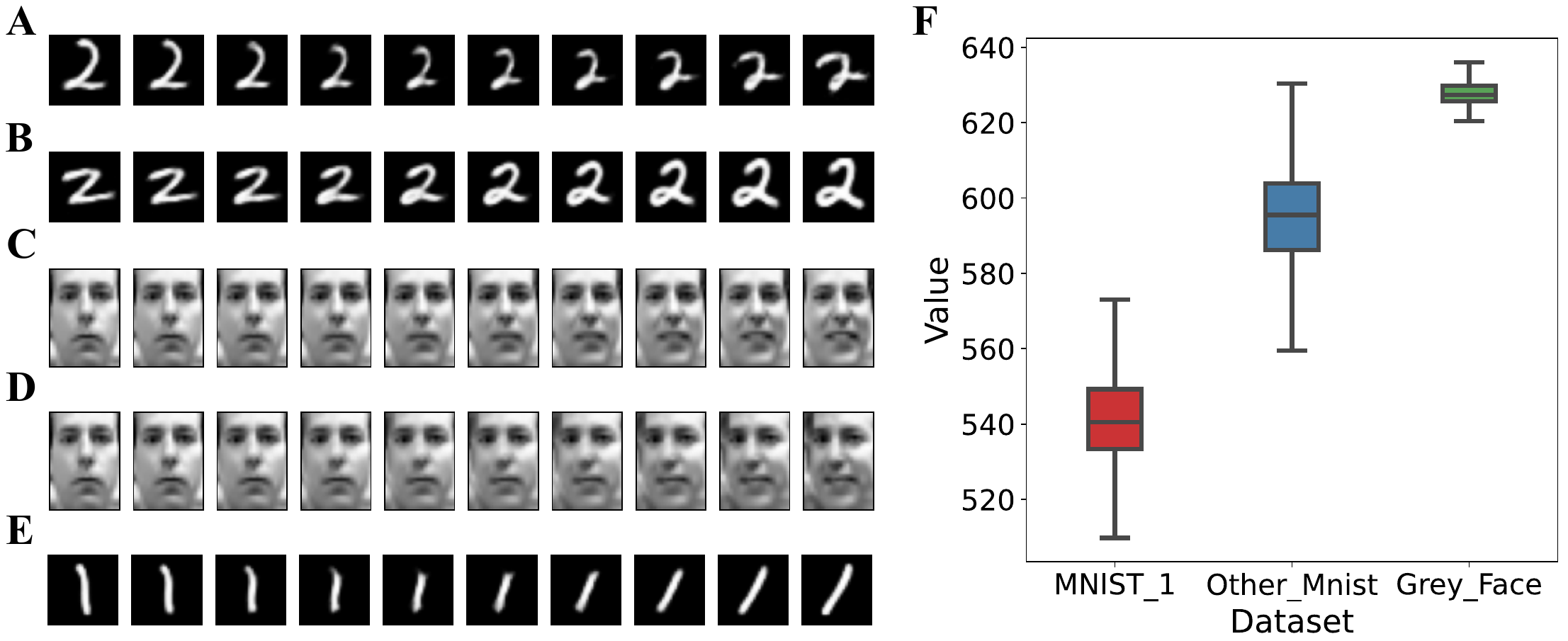}
	\caption{Results on image datasets. (\emph{A})-(\emph{E}) The results of generating images through linear interpolation in the character space using the handwritten ``2" from the MNIST dataset (\emph{A}), (\emph{B}), the "Gray Face" dataset (\emph{C}), (\emph{D}), and the handwritten ``1" from the augmented MNIST dataset (\emph{E}) as the training datasets. In each image, the leftmost and rightmost subplots are selected for interpolation. (\emph{F}) The distribution of $\tilde{z}_1^2+\tilde{z}_2^2+\cdots+\tilde{z}_{771}^2$ after inputting each digit into the autoencoder trained on the augmented MNIST dataset with handwritten ``1".}\label{fig8}
\end{figure}

After the first round of training, we select two images from these images, labeled as $X$ and $Y$ (the leftmost and rightmost images in {Fig.\ref{fig8}A and \ref{fig8}B}), with values $G(X)$ and $G(Y)$ in the character space, respectively. Then, we calculate $\lambda G(X)+(1-\lambda)G(Y)$, $\lambda \in (0,1)$ in the character space to obtain a series of linear interpolation points between $G(X)$ and $G(Y)$. By reconstructing these interpolation points using the decoder $\hat{G}\big(\lambda G(X)+(1-\lambda)G(Y)\big)$, we obtain a series of images between $X$ and $Y$. It can be seen that the transformation between these images is continuous and gradient, indicating that the GAMLA has learned the manifold structure in the high dimensional original space.

Furthermore, we conduct the same experiments on a more complex ``\href{https://cs.nyu.edu/~roweis/data.html}{Gray Face}" image dataset. This dataset contains $1965$ images with a pixel size of $20 \times 28$. Similarly, we first estimate the intrinsic dimension of the underlying manifold to be approximately $24$ (as shown in {Fig.\ref{fig7}B}). Then we train the GAMLA with a bottleneck layer of size $24$ and generate a series of linear interpolation in the character space analytically for the
leftmost and rightmost images in Fig.\ref{fig8}C and \ref{fig8}D.  The results also validate  that the images generated by interpolation represent a continuous transformation between the two selected images.

It is important to note that the above steps are accomplished with only the first round of training. After the second round of training with added nodes, we can further distinguish different patterns of images. According to the analysis above, the nonzero value of complementary representation $\tilde{\bm{z}}$ actually reflect the distance between a point from the manifold. In this case, an image of $28\times 28$ pixels falls in a $784$ dimensional space while we use the character space of around $13$. Thus the dimension of the complementary space reaches $771$, and the value of the $\|\tilde{\bm{z}}\|_2^2$ reflect the distance between a specific image to the reconstructed manifold. In Fig.\ref{fig8}F, we use the digit ``1" in MNIST as the training data and depict the distribution of the final $\|\tilde{\bm{z}}\|^2_2$ for three types of image patterns: digit ``1",  other digits, and the Gray Face images. The lowest value of $\|\tilde{\bm{z}}\|_2^2$ appears on digit ``1" , and the largest value of $\|\tilde{\bm{z}}\|^2_2$ appears on the Gray Face image set while the value of $\|\tilde{\bm{z}}\|^2_2$ for other digits in MNIST lies between digit ``1" and the face dataset. These results confirm the theoretical conclusions for GAMLA, showing that the value of the complementary representation $\tilde{\bm{z}}$ measures the distance between different patterns, even though strict zero values are seldom observed in real applications.

\section{Discussions}
In this section, several issues closely related to the mechanism and performance of GAMLA will be discussed.
\subsection{The Influence of the Network Structure and Noise}
\begin{figure}[h]
	\centering
	\includegraphics[width=0.99\textwidth]{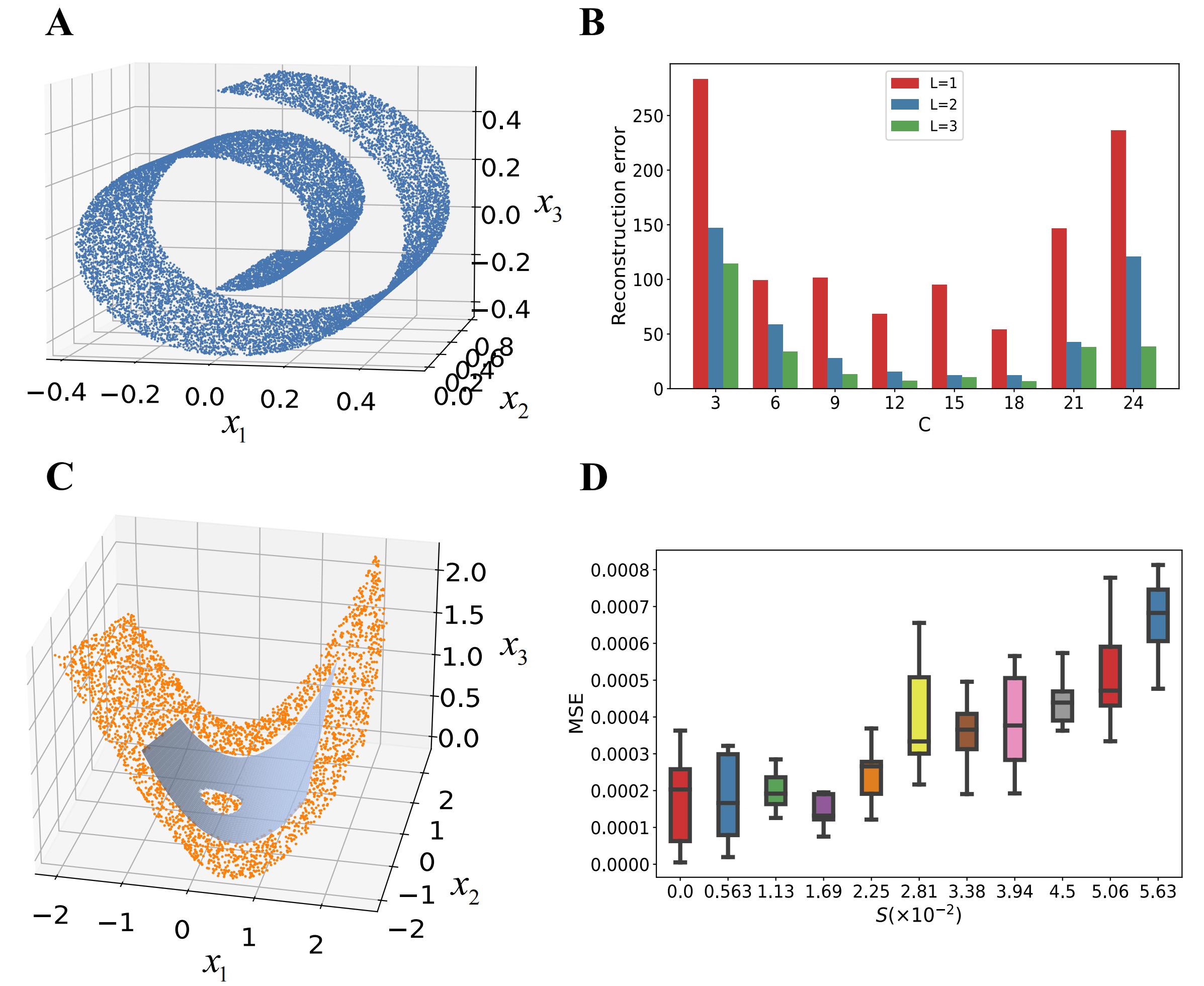}
	\caption{The impact of network structure and noise. (\emph{A}) Scatter plot of the Swiss Roll dataset. (\emph{B}) The distribution of the average sum of reconstruction errors for the Swiss Roll manifold trained with an autoencoder of structure $(L,C)$ over $10$ runs, where $(L,C)$ indicates that there are $2L+1$ hidden layers, and the dimensions of all hidden layers, except for the bottleneck layer, are $C$. (\emph{C}) The gray surface is the target manifold and the orange data points are not on the target manifold but satisfy the equation $R_4(x_1,x_2,x_3) \approx 0$. (\emph{D}) The distribution of the mean squared reconstruction errors of data points on the manifold when different levels of noise are added to the Swiss Roll manifold dataset as the training set.}\label{fig9}
\end{figure}

The performance of GAMLA, like any machine learning framework, is influenced by several hyperparameters and data conditions. The network structure—defined by the number of layers and the dimension of each layer, which correspond to the depth and width of GAMLA—is one of the most critical factors.
To analyze this impact, we use the Swiss Roll manifold as a benchmark and sample a data cloud with $20000$ data points, as shown in Fig.\ref{fig9}A. In this analysis, we examine GAMLA architectures composed of $2L+1$ fully connected hidden layers, each with dimension $C$ and utilizing the $\tanh$ activation function except for the bottleneck layer. We refer to such a model structure as $(L,C)$.

The average total reconstruction errors after 10 runs for different network structures are shown in Fig.\ref{fig9}B. These results demonstrate that both the number of layers $L$ and the dimension $C$ of each layer significantly affect the manifold reconstruction process. Interestingly, the error curve as a function of layer dimension
$C$ exhibits a U-shape. This is consistent with common machine learning principles: a layer dimension that is too small fails to provide sufficient complexity for the learning task, while an overly large dimension requires more training data, leading to increased difficulty in convergence \citep{Cybenko1989ApproximationBS,Lu2017TheEP,Neyshabur2014InSO}. Thus, there exists an optimal layer dimension $C$ that balances complexity and learnability.

Additionally, noise in real-world datasets increases the complexity of manifold learning, which can affect model performance\citep{annurev:/content/journals/10.1146/annurev-statistics-040522-115238}.
To examine the robustness of GAMLA against noise, we carry out noise tests for the benchmark dataset. Based on the results in Fig.\ref{fig9}B, we select an autoencoder architecture with a structure of $(3,18)$. Gaussian noises with standard deviation $\sigma$ are added to the Swill Roll data cloud and under different noise intensity $S$ we train the autoencoder with the noisy dataset $10$ times, with each training run comprising $500$ epochs.  The mean reconstruction error after each training is shown in Fig.\ref{fig9}D, implying that the used GAMLA framework performs well with noise strength $S$ under $2.25 \times 10^{-2}$.

\subsection{Theoretical Extensions}
The major theoretical result of this work is summarized in Proposition  \ref{prop1} which reveals that after the second round of training, the complementary representation $\tilde{\bm{z}} = R(\bm{x})$ indicates whether a point is located within the manifold, i.e., $R(\bm{x})=\bm{0}$ holds for any $\bm{x}\in \mathcal{M}$. As a matter of fact, several natural extensions of this conclusion arise.

First, it is important to acknowledge that in both the first and second rounds of training, the training sets contain only a limited number of samples from the manifold. Therefore, the learning process, which leverages strong generalization capabilities, results in $R(\bm{x})=0$ describing not only the manifold $\mathcal{M}$, but also potentially a larger manifold $\widetilde{\mathcal{M}} \subset \mathbb{R}^n$, of which $\mathcal{M}$ is a submanifold. This implies that for any $\bm{x} \in \widetilde{\mathcal{M}} \setminus \mathcal{M}$, we still have $R(\bm{x}) = 0$. To demonstrate this, we still consider the quadratic surface as expressed in Eq. \ref{manifold1} and the same training data set except a hole within it, as depicted in Fig.\ref{fig9}C. Leveraging the general function $R_4(x_1,x_2,x_3)=0$ learned by GAMLA, we generate $500000$ uniformly distributed data points within an expanded region and subsequently filter them to retain only those satisfying $|R_4(x_1,x_2,x_3)|<0.001$. As shown in Fig.\ref{fig9}C, the gray surface is the target manifold $\mathcal{M}$, while the orange points represent those that satisfy this condition but do not lie on $\mathcal{M}$. Intuitively, These orange points form a natural extension of $\mathcal{M}$, capturing both interior interpolations and outward extensions of the manifold. This characteristic suggests that GAMLA is capable of describing not just the original manifold but also its geometric structure in a more expansive sense. By learning these extensions, GAMLA can generate new points both inside and outside the manifold, effectively interpolating between known data and predicting the behavior of regions beyond the observed dataset.

On the other hand, Proposition \ref{prop1} ensures that  $R(\bm{x})\neq 0$ implies $\bm{x}\notin\mathcal{M}$. Therefore, in applications significantly nonzero values of $R(\bm{x})$ can serve as an indicator for points outside the manifold, which is especially useful for anomaly detection and explanation.

Anomaly explanation has become a critical topic, with many advances reported, particularly within the deep learning framework \citep{Yepmo2021AnomalyEA,Tritscher2023FeatureRX}. {However, existing anomaly detection methods are often difficult to interpret due to the lack of transparency in feature importance explanations \citep{Antwarg2021ExplainingAD, Tritscher2023FeatureRX}.} Unlike deep learning models, GAMLA goes beyond identifying abstract features by establishing relationships between normal sample features that are captured by $R(\bm{x})=0$. Particularly, the value of $R(\bm{x})$ provides insights into the type and degree of anomaly. As an example, in Fig.\ref{fig6}A and \ref{fig6}B, $P_1$ and $P_2$ are two outliers relative to the Swiss Roll manifold. Despite their proximity, their values of the complementary representation $\tilde{\bm{z}}=R(\bm{x})$ have opposite signs, indicating that one point deviates inward and the other outward from the manifold.
Additionally, it is noted that $R(\bm{x})=\rm{0}\in\mathbb{R}^{n-m}$ actually includes $n-m$ independent relations, allowing for a deeper understanding of which specific feature relationships are violated by anomalous points, i.e., $\tilde{z}_i \neq 0$ signifies that the data point violates the $i$th relationship. Therefore, we can further cluster the outliers based on their values of $R(\bm{x})$ according to the reasons for their anomalies.

\subsection{Learning of Complex Manifolds}
{Although GAMLA has shown effective learning capabilities for compact and bounded manifolds, its application to learning tasks on more complex manifolds is worthy of further investigation. For compact, boundaryless manifolds such as spheres and torus, which cannot be fully covered by a single coordinate chart \citep{annurev:/content/journals/10.1146/annurev-statistics-040522-115238}, we can initially segment these manifolds into two regions with each of them covered by an individual chart, and then employ GAMLA  to learn the structure of each segmented region respectively. In addition, prior research has converted periodic single-cell RNA-seq data into a learning problem framed within the context of circular manifolds. This transformation is accomplished by integrating periodic activation functions into the autoencoder, thereby enhancing its ability to learn circular manifold structures \citep{Liang2019LatentPP,Riba2021CellCG}. This methodology offers valuable insights for refining GAMLA in the study of compact, boundaryless manifolds. However, effectively learning the structural information of compact non-orientable manifolds, such as the Möbius strip, remains a challenging problem.}

\subsubsection*{Acknowledgments}
This work was supported by the National Natural Science Foundation of China (12471467), the National Key Research and Development Program of China (2018YFA0801103), the National Natural Science Foundation of China (12071330) to Ling Yang, the National Natural Science Foundation of China (12171350) to Huan-Fei Ma.

\bibliography{ref}
\bibliographystyle{iclr2023_conference}

\end{document}